\pdfoutput=1

\documentclass[11pt]{article}

\usepackage[preprint]{coling}
\usepackage{xcolor}

\usepackage{times}
\usepackage{latexsym}
\usepackage{multirow}

\usepackage[T1]{fontenc}
\usepackage[utf8]{inputenc}

\usepackage{microtype}
\usepackage{listings}
\usepackage{inconsolata}

\usepackage{graphicx}
\usepackage{svg}
\usepackage{mdframed}
\usepackage{float}

\title{Text Is Not All You Need: Multimodal Prompting Helps LLMs Understand Humor}

\author{Ashwin Baluja \\
  Northwestern University \\ 
  \texttt{baluja@u.northwestern.edu}}

\begin{document}
\maketitle
\begin{abstract}
While Large Language Models (LLMs) have demonstrated impressive natural language understanding capabilities across various text-based tasks,  understanding humor has remained a persistent challenge. Humor is frequently multimodal, relying on phonetic ambiguity, rhythm and timing to convey meaning. In this study, we explore a simple multimodal prompting approach to humor understanding and explanation. We present an LLM with both the text and the spoken form of a joke, generated using an off-the-shelf text-to-speech (TTS) system. Using multimodal cues improves the explanations of humor compared to textual prompts across all tested datasets.
\end{abstract}

\section{Introduction}
Despite remarkable advances in Natural Language Processing, particularly with Large Language Models (LLMs), the computational understanding of humor remains an elusive goal. Humor operates on multiple levels simultaneously, drawing on cultural context, current events, common sense, phonetic nuances, and rhythm to evoke a comedic response \cite{ambiguity, timing, subversion}. Recent studies have focused on analyzing the performance of large language models for understanding cultural norms \cite{mmlu2,mmlu1}, knowledge of current events \cite{livebench}, and common sense reasoning \cite{hellaswag,performance}. Yet, the unique challenge posed by computational humor, requiring a combination of all these tasks and information often conveyed through audio, has received comparatively little attention.

A fundamental aspect of verbal humor, particularly evident in puns, lies in linguistic ambiguity. Puns rely on homographs (words that are spelled identically with different meanings) and heterographs (words that are spelled differently but pronounced the same) \cite{semeval7}. Traditional text-based LLMs, constrained by their token-based processing architecture, struggle to capture these subtle linguistic features that yield essential clues into understanding a joke's underlying mechanics.

Our approach builds on prior research into humor understanding abilities present in LLMs \cite{reword} and demonstrates significant improvements over baseline textual prompting strategies for humor explanation. Our analysis examines both macro-level performance across humor datasets and micro-level effects through investigation of the model's internal representations and effects of text-to-speech (TTS) parameters.

\begin{figure*}[t]
\centering
\includegraphics[width=0.8\linewidth]{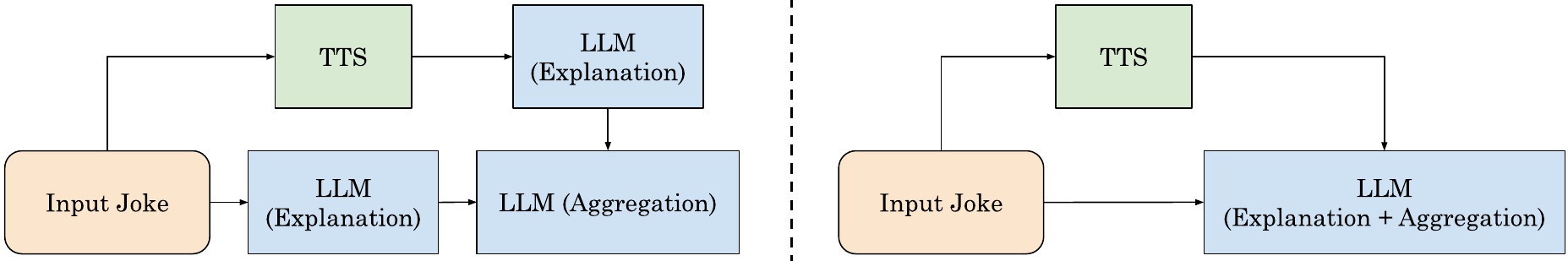}
  \caption {Multimodal prompting strategy overview. Left - separate text and audio explanations are generated, then aggregated. Right - combined text and audio are processed together for a single explanation. Aggregation refers to prompting intended to ensure a coherent output that does not represent the multiple input modalities.
  }
  \label{fig:method}
\end{figure*}

\section{Related Work}
\textbf{LLM-based humor classification.}
In \cite{finetuning}, pre-trained language models, from early BERT-like models to modern LLMs like LLaMa-3, were fine-tuned or prompted using Chain-of-Thought (CoT) and few-shot strategies for humor classification. Similarly, in \cite{reword}, CoT and few-shot example prompting were used for punchline detection and humor explanation. In both, humor specific examples for fine-tuning proved beneficial for humor understanding.

\noindent \textbf{Fused multimodality features.}
Most directly related to our work, several studies have developed multimodal features for humor detection ~\cite{neuralapproach,fused2}. Each has improved the performance of their task by incorporating non-text information into a fused representation, validating the importance of other multiple modalities. In contrast to our approach, these studies were conducted using BERT-like models and required training from scratch. We propose a simpler approach that does not require training and is fully compatible with pre-trained LLMs. 

\noindent \textbf{Training on paired modality datasets.}
In domains outside of humor, copious research has been conducted using datasets with 
multiple representations of the same items. These have been critical to the creation of multimodal language models. For example, LLaVa, a vision$+$language model, used detailed captions of images to generate  question and response pairs from a language-only LLM in order to train a vision-adapter \cite{liu2023llava}. 

\section{Methods}
We propose using multimodal prompts and audio synthesis to improve a model’s ability to capture the phonetic elements essential to understanding humor, specifically puns. The framework,  shown in Figure \ref{fig:method}, has two key components: generating audio from text and a prompting strategy that combines both the audio and text into a single prompt.

\subsection{Audio generation}
The most effective comedians rely on not only their material but also on carefully controlled and exaggerated cadence changes, volume and tonality of their words, as well as myriad more nuanced features. For replicability and breadth of trials, we use a simple, reproducible approach: text-based jokes are first converted into audio using \textbf{OpenAI's tts-1-hd}, an off-the-shelf text-to-speech (TTS) model. This procedure is broadly applicable and does not require existing audio datasets or the collection of human speech.  This method will be directly compared to only using text-based prompting on a diverse set of large data sets. Note that no additional ground-truth information (e.g., emphasis or timing), is provided to the LLM.

\subsection{Prompt configuration}
\begin{figure}[t]
\centering
\includegraphics[width=0.8\linewidth]{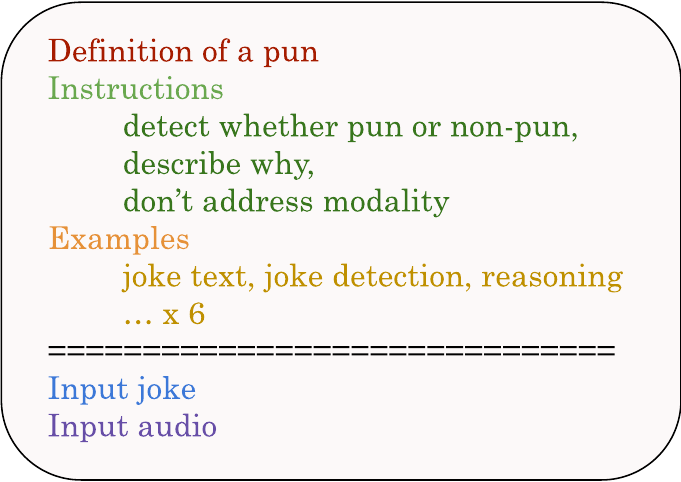}
  \caption {Multimodal prompting strategy where the LLM has both audio and text passed in at once.}
  \label{fig:prompt}
\end{figure}

Each prompt is composed of  general task definitions, chain-of-thought reasoning prompting, examples (for few-shot, in-context learning), and both input modalities (Fig \ref{fig:prompt}).

\noindent \textbf{Definitions and Instructions:} Each prompt begins with a concise definition of puns versus non-puns, accompanied by instructions for humor detection. The instructions request that the model identify whether the input is a pun or not. We explicitly \emph{did not} ask for an explanation at this point; \cite{reword} found this helps reduce hallucinated evidence of non-existent puns.

\noindent \textbf{Few-Shot Examples and chain-of-thought:} 
Each prompt includes  examples; using examples strongly benefit task performance \cite{fewshotlearners}. As suggest by \cite{reword}, each example includes chain-of-thought reasoning along with the detection result. In total, each prompt included six examples of pun explanation pairs, including both homographic and heterographic puns, selected from each dataset tested. Including few-shot examples also ensured a consistent output tone, making the results more directly comparable with ground-truth human provided explanations.

To obtain an \emph{explanation} of what makes the joke funny and why, we use prompts that encourage chain-of-thought reasoning~\cite{reword}. In this configuration, the reasoning for why a pun was detected as a pun or non-pun is used directly as the explanation. This format guides the model to accurately understand the task while avoiding biasing it towards interpreting the input as a pun. 

\noindent \textbf{Multimodal Aggregation:} In our multimodal setup, both text and audio are provided to the LLM.  Two approaches were tested. First, to mitigate the chances of the LLM  exclusively using either the text or the audio, we ran two parallel explanation processes. Each only had access to a single modality. Then, an aggregation prompting step was run, combining the two outputs into the final,  single, output (Figure \ref{fig:method}-Left).

Second, we provided both the audio and text to the model within a single prompt (Figure \ref{fig:method}-Right). In both setups, the prompt was carefully crafted to instruct the LLM to actively \emph{avoid discussing the source modality that it used} to answer. This was required as the target explanations in the datasets do not have any reference to modality (as they only had text). We found that the latter method outperformed the first; it will be used going forward. More details are provided in the ablation section.

\section{Experimental Setup}
\subsection{Datasets}
\label{ssec:datasets}
We evaluate our multimodal prompting approach using three datasets.

\noindent \textbf{SemEval 2017 Task 7}: This set contains 810 \& 647 puns (homographic \& heterographic), and 1077 non-puns.  It contains \underline{human annotations}:  noting the  pun-word  and the spelling or definition of the pun-word in both interpretations of the pun. Additionally, this set contains \underline{human explanations:} each pun has a human-provided, sentence-form explanation \cite{semeval7}. 

\noindent \textbf{Context-Situated Puns:} This dataset 
consists of 821 \& 1739 puns (homographic \& heterographic) with \underline{human annotations}~\cite{contextsituatedpuns}.  It has the same pun-word and
spelling/definition annotations as SemEval.  It does not contain  \underline{human explanations}.

\noindent \textbf{ExplainTheJoke}: This 
broad collection of jokes was scraped from the \texttt{ExplainTheJoke.com} website, containing 350 jokes. This dataset does not have \underline{human annotations}. Instead, each entry has \underline{human explanations} in paragraph-form, which themselves have high variability in style, length, quality and accuracy~\cite{explainthejoke}.

\subsection{Models}
For generating explanations, we utilized Gemini-1.5-Flash \cite{gemini}. At the time of the study, only the Gemini family of models offered API-based audio input, and Gemini-1.5-Flash offered the best balance between performance and affordability. This allowed us to effectively leverage the multimodal prompts central to our study, ensuring that the model could process auditory cues alongside textual data.

To avoid human biases in evaluating the quality of the generated free-form text explanations, a separate LLM  was used. Recently, using LLMs as \emph{judges} have been assessed favorably \cite{evaluation1}. Further, the presence of structured annotations for both pun datasets gave the judge ground-truth, so that it did not rely on its own understanding of the joke; details are provided in the next section. 

We chose GPT-4o \cite{gpt4o} as the judge.  We did not use Gemini-1.5-Flash or Gemini-1.5-Pro to avoid  potential biases from using the same model family for  generation and evaluation. Additionally, at the time of our research, GPT-4o was the strongest available model~\cite{lmarena}.\footnote{The final score was based on pairwise comparisons of each test sample. However, LLM-as-a-judge has been found to have strong positional bias, \emph{a priori} preferring the first element in each pair. To account for this, we run pairwise comparison twice, swapping the order of each pair. Final win rates are determined by averaging swapped and un-swapped win rates \cite{evaluation1,evaluation2}.}

\section{Results}
We present our results on the three datasets.  Interestingly, there is a wide disparity in the \emph{understanding} and  \emph{detection} of humor using LLMs.  While previous studies have shown close to saturated results in the simpler problem of \emph{detecting} the presence humor~\cite{reword}, an LLMs explanation of the humor is often incorrect.  Our results demonstrate the improvements in \emph{understanding} that are possible with multimodal inputs. 

\subsection{SemEval}
To evaluate our system, Gemini-1.5-Flash's generated explanation for each pun was paired with the human-provided explanations from the dataset. The judge, GPT-4o, was then asked to output whether explanation 1 was better than explanation 2, explanation 2 was better than explanation 1, or whether both explanations were of equal quality. The full judging prompt is included in Appendix \ref{ssec:appendixjudging}. This process was repeated twice - once for baseline textual prompting, and once for multimodal prompting, where the LLM was provided with both the text and audio. The judge was provided with the annotations each time, ensuring that the judgement considered the ground truth meaning. Win rate is reported as percent of times the model's explanation was preferred over the human's.

\begin{table}[h]
\centering
\begin{tabular}{c||c c|c c}

& \multicolumn{2}{c}{Heterograph} & \multicolumn{2}{c}{Homograph} \\ 
\hline
& Win \% & Tie \% & Win \% & Tie \%\\
\hline
Baseline & 47.76 & 5.64 & 68.89 & 8.40 \\
\textbf{with audio} & \textbf{51.74} & 4.56 & \textbf{72.59} & 6.36 \\
\end{tabular}
\caption{Results for SemEval comparing baseline and multimodal prompting vs. human explanations.}
\label{table:semeval}
\end{table}

Table \ref{table:semeval} shows that incorporating audio significantly improves performance over baseline across both homographs and heterographs. Performance increased in both by approximately 4\%.

\subsection{Context-Situated Puns}
Unlike the previous dataset, with no human explanations available, we cannot compare each LLM output to a human baseline.  Instead, we compare the LLM outputs (created with and without audio input) directly with each other.  To ensure that the judge-LLM is given the correct context, the annotations that were provided in the dataset are also given as input. Here, win-rate is reported as the percent of times the result of one prompting strategy was preferred over the other. 

\begin{table}[h]
\centering
\begin{tabular}{c||c c|c c}
& \multicolumn{2}{c}{Heterograph} & \multicolumn{2}{c}{Homograph} \\ 
\hline
& Win \% & Tie \% & Win \% & Tie \% \\
\hline
Baseline            & 33.87 & \multirow{2}{*}{29.65} & 35.08 & \multirow{2}{*}{28.08} \\
\textbf{with audio} & \textbf{36.49} & & \textbf{36.85} &\\ 
\end{tabular}
\caption{Results for Context-Situated Puns dataset comparing baseline vs multimodal prompting.}
\label{table:context_puns}
\end{table}

As shown in Table \ref{table:context_puns}, the addition of audio cues again provided improvements in both homographic and heterographic cases.

\subsection{ExplainTheJoke}
Here, we evaluated the model's ability to generate detailed joke explanations in domains outside of puns. As this dataset lacked detailed annotations and only included inconsistent-quality explanations, we attempted to generate a more normalized explanation by first asking GPT to summarize the provided human explanations. This summary was then used as the relevant context to the judge-LLM. The remainder of the evaluation proceeded in the same manner as the Context-Situated Puns; we performed pairwise comparison directly between the results of baseline and multimodal prompting. 

\begin{table}[h]
\centering
\begin{tabular}{c||c c}
& Win \% & Tie \% \\
\hline
Baseline & 12.81 & \multirow{2}{*}{71.75} \\
\textbf{with audio} & \textbf{15.44} & \\
\end{tabular}
\caption{Results for ExplainTheJoke dataset comparing baseline vs multimodal prompting.}
\label{table:explainthejoke}
\end{table}

Table \ref{table:explainthejoke}  reveals that even with jokes that are not puns, using the mutli-modal prompting improves performance. While phonetic ambiguity likely explains many of the performance gains for the previously studied datasets, these results suggest that other more nuanced effects are successfully conveyed by including audio. 

\subsection{Analysis}
We conducted three key analyses to assess the proposed multimodal prompting. First, in an ablation study, we tested various details of multimodal prompting, finding that pure audio-only prompting performs far worse than pure textual prompting (Appendix \ref{sec:appendixablations}). Second, we analyzed whether incorporating audio genuinely preserved phonetic ambiguity. Through a detailed examination of the logits of an LLM transcribing puns, we observed that the model assigned significant probability to both potential spellings of the pun-word (\emph{e.g.}, “weight” vs. “wait”), indicating that the phonetic cues \emph{were} captured in its internal representations (Appendix \ref{sec:appendixlogits}). Lastly, we explored sensitivity to voice parameters but found no significant evidence that variations in voice type systematically affected results (Appendix \ref{sec:appendixvoices}).

\section{Conclusion and Future Work}
In this study, we demonstrated that incorporating auditory cues into multimodal prompts significantly improves Large Language Models' ability to understand and explain humor, particularly in cases involving phonetic ambiguity. Our approach, leveraging readily available APIs and open-source models, offers a straightforward yet effective enhancement to existing LLM capabilities.

There are several avenues for extending this research. A deeper study into the effects of voice characteristics on humor interpretation could reveal how tone, pitch, or speaker identity affects comedic understanding. Although our TTS-based approach was effective, it did not capture nuances like timing and rhythm; similarly, integrating video input could convey facial expressions and other cues essential for humor. Finally, expanding our approach to include different forms of humor and other domains where multimodal inputs play a crucial role will further broaden the capabilities of LLMs in understanding the subtleties of user queries.

\section{Limitations}
\textbf{Prompt Sensitivity:} The success of our multimodal prompting approach depends heavily on prompt design. LLMs, particularly those using the Gemini architecture, are highly sensitive to the phrasing and structure of prompts. Small variations can lead to significant differences in output quality, necessitating extensive tuning to optimize performance. This reliance on precise prompt crafting limits the scalability and generalization of the approach to new tasks.

\noindent\textbf{Nuances Beyond Phonetic Ambiguity:} While our method demonstrated improved understanding of phonetic ambiguity in humor, it falls short in capturing more nuanced comedic elements such as timing, cadence, and rhythm. Our TTS-based approach does not fully convey the subtleties of real human speech, which are critical for interpreting humor beyond wordplay. This limitation suggests the need for richer audio models or the integration of additional modalities, such as video, to capture non-verbal cues.

\noindent\textbf{Evaluation Challenges:} Our evaluation relied on automated LLM-based judging, which, while efficient, may not fully capture the nuanced quality of humor explanations. Future studies should incorporate more robust evaluation strategies, such as human assessments, or using stronger models as they are released, to better gauge the effectiveness of these approaches in real-world scenarios.

\section{Ethics Statement}
Large Large Language Models can produce offensive and incorrect statements. In the process of explaining comedy, they may frequently encounter harmful stereotypes and offensive content. Both correct and incorrect explanations can result in an LLM outputting potentially hurtful answers. It is advisable for users to exercise caution and avoid prompting the LLM with potentially offensive jokes, so as to avoid the output perpetuating incorrect stereotypes. This work is released with the intent of research purposes only.  

\bibliography{custom}

\begin{thebibliography}{23}
\providecommand{\natexlab}[1]{#1}

\bibitem[{Aggarwal et~al.(2023)Aggarwal, Pandey, and Vishwakarma}]{fused2}
Sajal Aggarwal, Ananya Pandey, and Dinesh~Kumar Vishwakarma. 2023.
\newblock \href {https://doi.org/10.1109/WCONF58270.2023.10235179} {Multimodal sarcasm recognition by fusing textual, visual and acoustic content via multi-headed attention for video dataset}.
\newblock In \emph{2023 World Conference on Communication \& Computing (WCONF)}, pages 1--5.

\bibitem[{Attardo and Pickering(2011)}]{timing}
Salvatore Attardo and Lucy Pickering. 2011.
\newblock \href {https://doi.org/doi:10.1515/HUMR.2011.015} {Timing in the performance of jokes}.
\newblock \emph{HUMOR}, 24(2):233--250.

\bibitem[{Brown et~al.(2020)Brown, Mann, Ryder, Subbiah, Kaplan, Dhariwal, Neelakantan, Shyam, Sastry, Askell, Agarwal, Herbert-Voss, Krueger, Henighan, Child, Ramesh, Ziegler, Wu, Winter, Hesse, Chen, Sigler, Litwin, Gray, Chess, Clark, Berner, McCandlish, Radford, Sutskever, and Amodei}]{fewshotlearners}
Tom Brown, Benjamin Mann, Nick Ryder, Melanie Subbiah, Jared~D Kaplan, Prafulla Dhariwal, Arvind Neelakantan, Pranav Shyam, Girish Sastry, Amanda Askell, Sandhini Agarwal, Ariel Herbert-Voss, Gretchen Krueger, Tom Henighan, Rewon Child, Aditya Ramesh, Daniel Ziegler, Jeffrey Wu, Clemens Winter, Chris Hesse, Mark Chen, Eric Sigler, Mateusz Litwin, Scott Gray, Benjamin Chess, Jack Clark, Christopher Berner, Sam McCandlish, Alec Radford, Ilya Sutskever, and Dario Amodei. 2020.
\newblock \href {https://proceedings.neurips.cc/paper_files/paper/2020/file/1457c0d6bfcb4967418bfb8ac142f64a-Paper.pdf} {Language models are few-shot learners}.
\newblock In \emph{Advances in Neural Information Processing Systems}, volume~33, pages 1877--1901. Curran Associates, Inc.

\bibitem[{Bucaria(2004)}]{ambiguity}
Chiara Bucaria. 2004.
\newblock \href {https://doi.org/doi:10.1515/humr.2004.013} {Lexical and syntactic ambiguity as a source of humor: The case of newspaper headlines}.
\newblock \emph{HUMOR}, 17(3):279--309.

\bibitem[{Chiang et~al.(2024)Chiang, Zheng, Sheng, Angelopoulos, Li, Li, Zhang, Zhu, Jordan, Gonzalez, and Stoica}]{lmarena}
Wei-Lin Chiang, Lianmin Zheng, Ying Sheng, Anastasios~Nikolas Angelopoulos, Tianle Li, Dacheng Li, Hao Zhang, Banghua Zhu, Michael Jordan, Joseph~E. Gonzalez, and Ion Stoica. 2024.
\newblock \href {https://arxiv.org/abs/2403.04132} {Chatbot arena: An open platform for evaluating llms by human preference}.
\newblock \emph{Preprint}, arXiv:2403.04132.

\bibitem[{{{Gemini Team}}(2024)}]{gemini}
{{Gemini Team}}. 2024.
\newblock \href {https://arxiv.org/abs/2403.05530} {Gemini 1.5: Unlocking multimodal understanding across millions of tokens of context}.
\newblock \emph{Preprint}, arXiv:2403.05530.

\bibitem[{Hasan et~al.(2021)Hasan, Lee, Rahman, Zadeh, Mihalcea, Morency, and Hoque}]{neuralapproach}
Md~Kamrul Hasan, Sangwu Lee, Wasifur Rahman, Amir Zadeh, Rada Mihalcea, Louis-Philippe Morency, and Ehsan Hoque. 2021.
\newblock Humor knowledge enriched transformer for understanding multimodal humor.
\newblock In \emph{Proceedings of the AAAI conference on artificial intelligence}, volume~35, pages 12972--12980.

\bibitem[{Hendrycks et~al.(2021{\natexlab{a}})Hendrycks, Burns, Basart, Critch, Li, Song, and Steinhardt}]{mmlu2}
Dan Hendrycks, Collin Burns, Steven Basart, Andrew Critch, Jerry Li, Dawn Song, and Jacob Steinhardt. 2021{\natexlab{a}}.
\newblock Aligning ai with shared human values.
\newblock \emph{Proceedings of the International Conference on Learning Representations (ICLR)}.

\bibitem[{Hendrycks et~al.(2021{\natexlab{b}})Hendrycks, Burns, Basart, Zou, Mazeika, Song, and Steinhardt}]{mmlu1}
Dan Hendrycks, Collin Burns, Steven Basart, Andy Zou, Mantas Mazeika, Dawn Song, and Jacob Steinhardt. 2021{\natexlab{b}}.
\newblock Measuring massive multitask language understanding.
\newblock \emph{Proceedings of the International Conference on Learning Representations (ICLR)}.

\bibitem[{Hua(2024)}]{alm}
Chris Hua. 2024.
\newblock \href {https://tincans.ai/slm3} {Gazelle v0.2}.

\bibitem[{Liu et~al.(2023)Liu, Li, Wu, and Lee}]{liu2023llava}
Haotian Liu, Chunyuan Li, Qingyang Wu, and Yong~Jae Lee. 2023.
\newblock Visual instruction tuning.
\newblock In \emph{NeurIPS}.

\bibitem[{Miller et~al.(2017)Miller, Hempelmann, and Gurevych}]{semeval7}
Tristan Miller, Christian Hempelmann, and Iryna Gurevych. 2017.
\newblock \href {https://doi.org/10.18653/v1/S17-2005} {{S}em{E}val-2017 task 7: Detection and interpretation of {E}nglish puns}.
\newblock In \emph{Proceedings of the 11th International Workshop on Semantic Evaluation ({S}em{E}val-2017)}, pages 58--68, Vancouver, Canada. Association for Computational Linguistics.

\bibitem[{OpenAI(2024)}]{gpt4o}
OpenAI. 2024.
\newblock \href {https://arxiv.org/abs/2410.21276} {Gpt-4o system card}.
\newblock \emph{Preprint}, arXiv:2410.21276.

\bibitem[{Shao et~al.(2024)Shao, Basit, Karri, and Shafique}]{performance}
Minghao Shao, Abdul Basit, Ramesh Karri, and Muhammad Shafique. 2024.
\newblock \href {https://doi.org/10.1109/ACCESS.2024.3482107} {Survey of different large language model architectures: Trends, benchmarks, and challenges}.
\newblock \emph{IEEE Access}, pages 1--1.

\bibitem[{Sun et~al.(2022)Sun, Narayan-Chen, Oraby, Gao, Chung, Huang, Liu, and Peng}]{contextsituatedpuns}
Jiao Sun, Anjali Narayan-Chen, Shereen Oraby, Shuyang Gao, Tagyoung Chung, Jing Huang, Yang Liu, and Nanyun Peng. 2022.
\newblock \href {https://doi.org/10.18653/v1/2022.emnlp-main.306} {Context-situated pun generation}.
\newblock In \emph{Proceedings of the 2022 Conference on Empirical Methods in Natural Language Processing}, pages 4635--4648, Abu Dhabi, United Arab Emirates. Association for Computational Linguistics.

\bibitem[{theblackcat102()}]{explainthejoke}
theblackcat102.
\newblock \href {https://huggingface.co/datasets/theblackcat102/joke_explaination} {Theblackcat102/joke\_explaination - datasets at hugging face}.

\bibitem[{Wang et~al.(2024)Wang, Li, Chen, Cai, Zhu, Lin, Cao, Kong, Liu, Liu, and Sui}]{evaluation2}
Peiyi Wang, Lei Li, Liang Chen, Zefan Cai, Dawei Zhu, Binghuai Lin, Yunbo Cao, Lingpeng Kong, Qi~Liu, Tianyu Liu, and Zhifang Sui. 2024.
\newblock \href {https://doi.org/10.18653/v1/2024.acl-long.511} {Large language models are not fair evaluators}.
\newblock In \emph{Proceedings of the 62nd Annual Meeting of the Association for Computational Linguistics (Volume 1: Long Papers)}, pages 9440--9450, Bangkok, Thailand. Association for Computational Linguistics.

\bibitem[{Warren et~al.(2021)Warren, Barsky, and McGraw}]{subversion}
Caleb Warren, Adam Barsky, and A.~Peter McGraw. 2021.
\newblock \href {https://doi.org/10.1177/1088868320961909} {What makes things funny? an integrative review of the antecedents of laughter and amusement}.
\newblock \emph{Personality and Social Psychology Review}, 25(1):41--65.
\newblock PMID: 33342368.

\bibitem[{White et~al.(2024)White, Dooley, Roberts, Pal, Feuer, Jain, Shwartz-Ziv, Jain, Saifullah, Naidu, Hegde, LeCun, Goldstein, Neiswanger, and Goldblum}]{livebench}
Colin White, Samuel Dooley, Manley Roberts, Arka Pal, Ben Feuer, Siddhartha Jain, Ravid Shwartz-Ziv, Neel Jain, Khalid Saifullah, Siddartha Naidu, Chinmay Hegde, Yann LeCun, Tom Goldstein, Willie Neiswanger, and Micah Goldblum. 2024.
\newblock \href {https://arxiv.org/abs/2406.19314} {Livebench: A challenging, contamination-free llm benchmark}.
\newblock \emph{Preprint}, arXiv:2406.19314.

\bibitem[{Wu et~al.(2024)Wu, Huang, and Lau}]{finetuning}
Shih-Hung Wu, Yu-Feng Huang, and Tsz-Yeung Lau. 2024.
\newblock Humour classification by fine-tuning llms: Cyut at clef 2024 joker lab subtask humour classification according to genre and technique.
\newblock In \emph{Working Notes of the Conference and Labs of the Evaluation Forum (CLEF 2024). CEUR Workshop Proceedings}, pages 1933--1947.

\bibitem[{Xu et~al.(2024)Xu, Yuan, Chen, and Yang}]{reword}
Zhijun Xu, Siyu Yuan, Lingjie Chen, and Deqing Yang. 2024.
\newblock \href {https://arxiv.org/abs/2404.13599} {"a good pun is its own reword": Can large language models understand puns?}
\newblock \emph{Preprint}, arXiv:2404.13599.

\bibitem[{Zellers et~al.(2019)Zellers, Holtzman, Bisk, Farhadi, and Choi}]{hellaswag}
Rowan Zellers, Ari Holtzman, Yonatan Bisk, Ali Farhadi, and Yejin Choi. 2019.
\newblock Hellaswag: Can a machine really finish your sentence?
\newblock In \emph{Proceedings of the 57th Annual Meeting of the Association for Computational Linguistics}.

\bibitem[{Zheng et~al.(2024)Zheng, Chiang, Sheng, Zhuang, Wu, Zhuang, Lin, Li, Li, Xing, Zhang, Gonzalez, and Stoica}]{evaluation1}
Lianmin Zheng, Wei-Lin Chiang, Ying Sheng, Siyuan Zhuang, Zhanghao Wu, Yonghao Zhuang, Zi~Lin, Zhuohan Li, Dacheng Li, Eric~P. Xing, Hao Zhang, Joseph~E. Gonzalez, and Ion Stoica. 2024.
\newblock Judging llm-as-a-judge with mt-bench and chatbot arena.
\newblock In \emph{Proceedings of the 37th International Conference on Neural Information Processing Systems}, NIPS '23, Red Hook, NY, USA. Curran Associates Inc.

\end{thebibliography}

\appendix

\section{Appendix}
\label{sec:appendix}

\subsection{Prompts}
\label{sec:appendixprompts}
These prompts are based on those used in \cite{reword}, modified for use with multimodal prompting

\subsubsection{Explanation}
The prompts used for generating explanations are shown in Figure \ref{fig:explanation}.

\label{ssec:appendixexplanation}

\subsubsection{Judging}
The prompts used for judging are shown in Figure~\ref{fig:judging}.

\label{ssec:appendixjudging}

\subsection{Phonetic Ambiguity}
\label{sec:appendixlogits}

In order to test whether phonetic ambiguity is preserved by including an audio version of the joke, we directly analyze the logits of a transcription task. Following the same pattern as in the joke explanation task, we convert the text to audio using OpenAI's tts-1-hd model. As publicly available LLM APIs do not provide logit outputs, we use \cite{alm}, an open-source model available on Hugging Face. 

As a simplified task, we converted the word, "Where," to audio, and asked the LLM to transcribe the file. If phonetic ambiguity is preserved, the model would output homophones for "where" as highly probably alternatives. As shown in \ref{fig:wherelogits}, this is the case: "wear," "ware," "here," and "there," are all present in the top ten highest probabilities.

\begin{figure}[H]
\includegraphics[width=\linewidth]{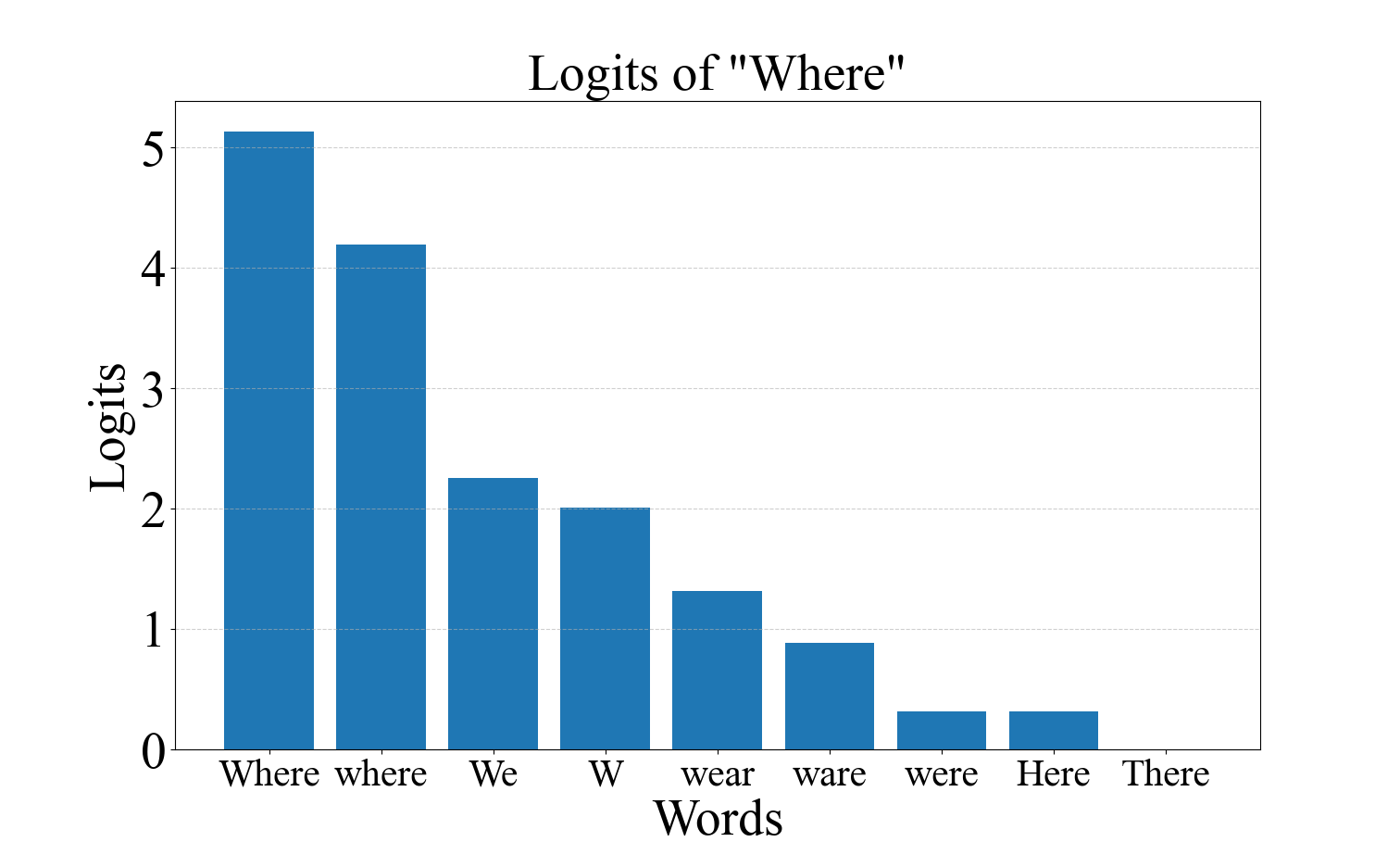}
\caption{Logits of transcribing an audio file containing the word, "Where".}
\end{figure}
\label{fig:wherelogits}

Further, we test the logits in a realistic pun scenario. We tested on the pun, "Patience is a heavy weight," where the pun-word is "weight," and the alternate spelling is "wait." As shown in \ref{fig:weightlogits}, "weight" and "wait" are the top two tokens with the highest probability. 

\begin{figure}[H]
\includegraphics[width=\linewidth]{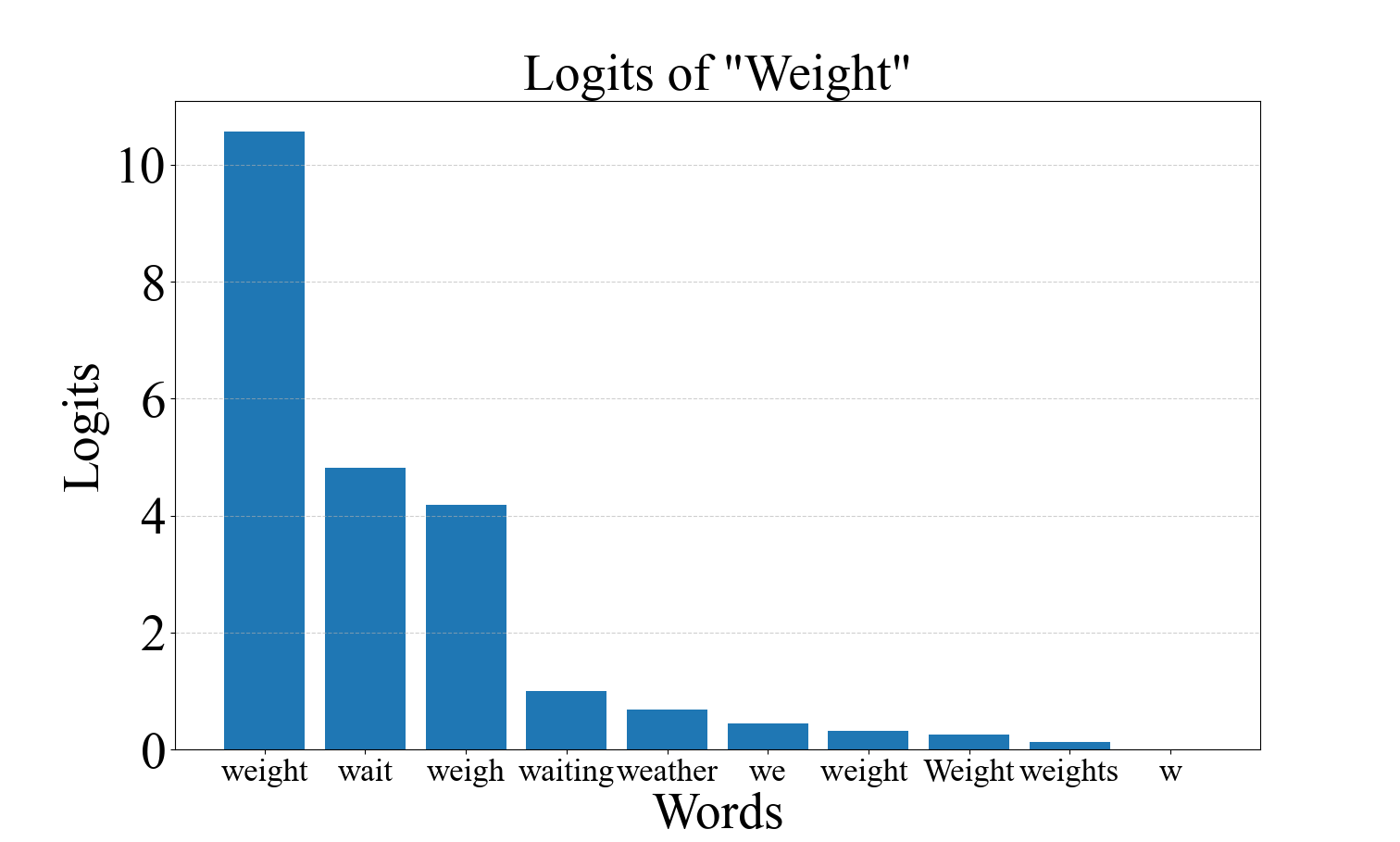}
\caption{Logits of transcribing an audio file containing a pun, focusing on the pun-word, "Weight".}
\end{figure}
\label{fig:weightlogits}

\subsection{Ablation Study}
\label{sec:appendixablations}
To assess the impact of various prompting configurations, we conducted an ablation study (Table \ref{table:ablation}) using audio-only prompting, and no wording to prevent addressing modality in the answer.  Results are shown in Table~\ref{table:ablation}.

\begin{table}[h]
\small
\centering
\begin{tabular}{c||c |c}
& Heterograph & Homograph \\ 
\hline
& Model Win \% & Model Win \% \\
\hline
No text, Audio Only & 25.50 & 55.86 \\
No aggregation wording & 48.61 & 71.05 \\
\hline
Our System & \textbf{51.74} & \textbf{72.59} \\
\end{tabular}
\caption{Ablation results comparing results for different prompting strategies.}
\label{table:ablation}
\end{table}

Additionally, we extensively tested how the separate prompting strategy shown in Figure~\ref{fig:method}-Left worked in comparison to the combined audio-text strategy employed.  In every test performed, the separate strategy was significantly worse than the combined strategy.

\vfill\eject

\subsection{Effects of Choice of Voice}
\label{sec:appendixvoices}

Comedic perception may be influenced by the actual voice of the comedian.   Table \ref{table:voices} presents a comparison of performance across different TTS voice types on the SemEval dataset using the full multimodal prompting strategy. The results show some variations in win rates, particularly for heterographs, but no clear pattern emerges that suggests a statistically significant advantage for any specific voice type. Although the androgynous voice (Alloy) slightly outperformed overall, the differences were not substantial enough to draw definitive conclusions regarding the impact of voice characteristics on the model's performance. Alloy was chosen for consistency throughout the experiments.  This remains open for future study.

\begin{table}[ht!]
\small
\centering
\begin{tabular}{c||c c|c c}
& \multicolumn{2}{c}{Heterograph} & \multicolumn{2}{c}{Homograph} \\ 
\hline
& Win \% & Tie \% & Win \% & Tie \% \\
\hline
Nova (female)& 44.59 & 5.25 & 71.48 & 6.30 \\
Onyx (male)& 45.44 & 5.65 & 73.33 & 6.42 \\
Alloy (androgynous) & \textbf{51.74} & 4.56 & 72.59 & 6.36 \\
Onyx + Alloy & 47.91 & 3.79 & \textbf{73.09} & 5.74 \\
\end{tabular}
\caption{Performance on SemEval for different OpenAI TTS voice types. Alloy is used throughout the experiments in this study. "Onyx + Alloy" had two different audio files, along with text, passed in to the LLM at once.}
\label{table:voices}
\end{table}

\begin{figure*}[h!]
\begin{mdframed}
{
\setlength{\parindent}{0pt}
<*Definition*>
Puns are a form of wordplay exploiting different meanings of a word or similar-sounding words, while non-puns are jokes or statements that don't rely on such linguistic ambiguities.

<*Instruction*>
Determine whether the given text and audio is a pun. The audio provided is spoken version of the input text. It is provided in 1 different voice(s). Please see if hearing the pun aloud helps you determine whether the text is a pun, and if so, why. Give your reasons first, then make your final decision clearly. You should either say "The text input is a pun" or say "The given text is a non-pun", despite the fact that you are given both text and audio. Do not reference the fact that you are given both text and audio. Only use the most likely explanation, taking into account information from both formats. You must output the current status in a parsable JSON format. An example output looks like:
{"Reason": "XXX", "Choice": "The given text is a XXX"}

<*Examples*>
Text: When the waiter told me they were out of corn I said , ' That really shucks . '
Output:
{"Reason": "The text plays on the double meaning of the word 'shucks'. 'Shucks' refers to both the act of removing the husk from corn and is a homophone for 'sucks', which is used colloquially to express disappointment.", "Choice": "The given text is a pun"}

Text: Desperate times call for desperate measures .
Output:
{"Reason": "The text is an idiomatic expression meaning that one may need to take drastic actions in difficult situations. It does not exploit different meanings of a word or similar-sounding words.", "Choice": "The given text is a non-pun"}

Text: A tangled bell ringer tolled himself off .
Output:
{"Reason": "The text plays on the homophones 'tolled' and 'told', using the word 'tolled' in the context of a bell ringer (which relates to the ringing or tolling of bells) and 'told' as in scolding oneself (told sb off). This creates a humorous double meaning.", "Choice": "The given text is a pun"}

Text: Don't bite the hand that feeds you .
Output:
{"Reason": "The text is an idiomatic expression meaning one should not act ungratefully towards those who provide for them. It does not rely on a play on words or different meanings of the same word.", "Choice": "The given text is a non-pun"}

Text: An illiterate fisherman was lost at c .
...
{"Reason": "The text is an idiomatic expression that suggests it's better to be cautious than to get into trouble. It does not rely on the ambiguity of words or similar-sounding words for a humorous effect.", "Choice": "The given text is a non-pun"}

<*Your Response*>
Text: Patience is a virtue heavy in wait
Audio: <audio>
Output:
}
\end{mdframed}
\caption{Explanation prompt for an LLM, including examples and an example input pun.}
\label{fig:explanation}
\end{figure*}

\begin{figure*}[h!]
\begin{mdframed}
{
\setlength{\parindent}{0pt}
<*Definition*>
Puns are a form of wordplay exploiting different meanings of a word or similar-sounding words.

<*Instruction*>
Below is a pun text, double meanings of the pun and two corresponding explanations. Please carefully judge which explanation is of better quality. Any explanation that fails to indicate the correct pun, misses the potential phonetic similarity between pun-alternative word pair, misses a layer of correct meaning in the pun or contains other errors is a worse explanation. Meanwhile, explanations without the above errors are better explanations. To complete the task, you must cautiously choose from one of the three answers: "Explanation 1 is much better", "Explanation 2 is much better", "Explanation 1 and 2 are of similar quality". Additionally, You must output the current status in a parsable JSON format. An example output looks like:
{"Choice": "XXX"}

<*Your Response*>
Text: Hockey players are always terrible chess players since they aren\'t handy.
Double Meanings of the Pun: 1. pun word and its meaning: handy <useful and convenient>. 2. alternative word and its meaning: hand <the (prehensile) extremity of the superior limb>.
Explanation 1: The text plays on the double meaning of \'handy\'. \'Handy\' can refer to being skilled with one\'s hands, which is relevant to hockey players, but it can also mean \'nearby\' or \'convenient\', which is relevant to chess players. The joke lies in the contrast between the two meanings, suggesting that hockey players are not good at chess because they are not \'handy\' in the sense of being close to the chessboard.
Explanation 2: The text plays on the double meaning of the word \'handy\'. \'Handy\' can refer to being skilled or useful, but in the context of hockey, it also refers to the use of one\'s hands, which is not allowed in chess.
Output:
}
\end{mdframed}
\caption{Judging prompt for an LLM, including an example annotation and two potential explanations.}
\label{fig:judging}
\end{figure*}

\end{document}